\documentclass[11pt,letterpaper]{article}
\usepackage{emnlp2016}
\usepackage{times}
\usepackage{latexsym}
\usepackage{array}

\emnlpfinalcopy

\def\emnlppaperid{420}
\usepackage{amssymb}
\usepackage{url}
\usepackage{algorithm}
\usepackage[noend]{algorithmic}
\usepackage{tikz}
\usetikzlibrary{shapes,arrows}
\algsetup{linenosize=\small}
\usepackage{graphicx}
\usepackage{caption}
\usepackage{subcaption}
\usepackage{amsmath}
\usepackage{CJK}
\def\namecite{\newcite}

\let\tempone\enumerate
\let\temptwo\endenumerate
\renewenvironment{enumerate}{\tempone
\setlength{\itemsep}{0.5em}
\setlength{\parskip}{0pt}
\setlength{\parsep}{0pt}}{\temptwo}

\title{Exploring phrase-compositionality in skip-gram models}

\author{Xiaochang Peng \and Daniel Gildea\\
  {\tt xpeng,gildea@cs.rochester.edu}}

\date{}

\begin{document}

\maketitle

\begin{abstract}
In this paper, we introduce a variation of the skip-gram model which jointly learns 
distributed word vector representations and their way of composing to form phrase embeddings.
In particular, we propose a learning procedure that incorporates a phrase-compositionality
function which can capture how we want to compose phrases vectors from their component
word vectors. Our experiments show 
improvement in word and phrase similarity tasks as well as syntactic tasks like dependency
parsing using the proposed joint models.
\end{abstract}

\section{Introduction}
\begin{figure}[!tbp]
    \scalebox{0.5}{
\begin{subfigure}[b]{0.4\textwidth}
        \includegraphics{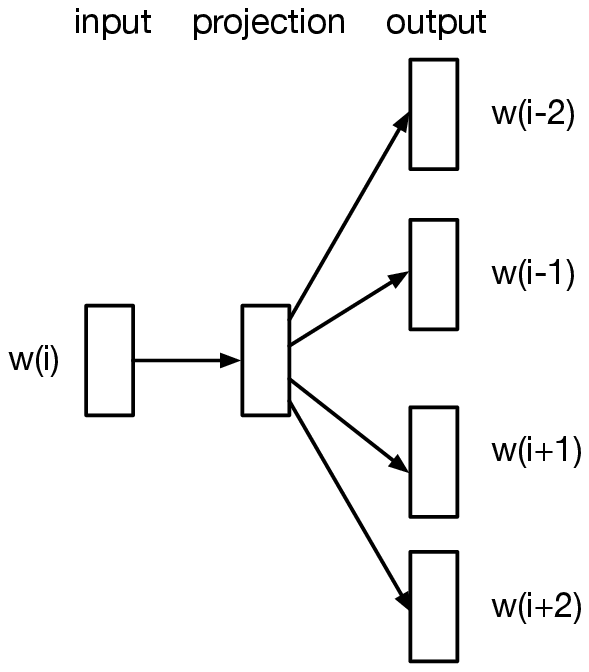}
    \caption{word-level skip-gram}
    \label{fig:1a}
\end{subfigure}
\hfill
\begin{subfigure}[b]{0.55\textwidth}
    \includegraphics{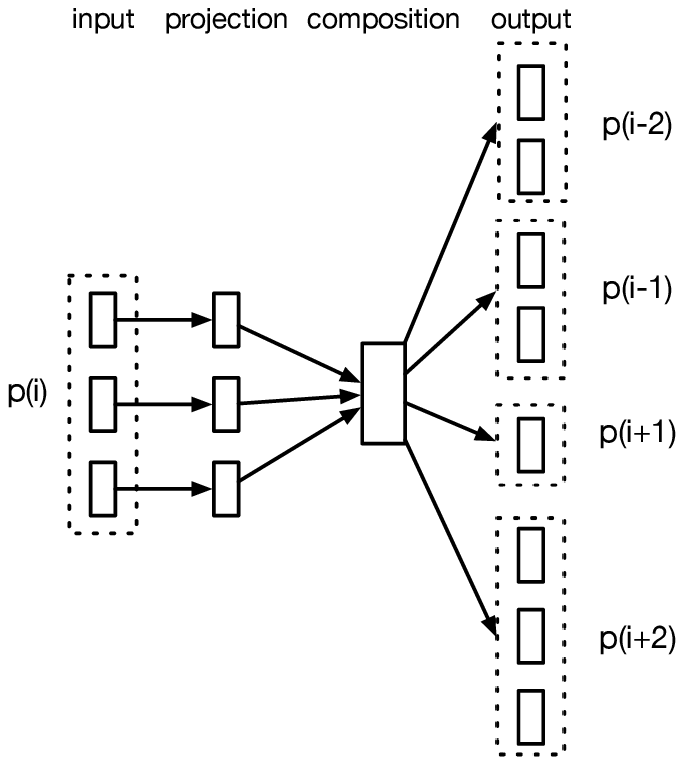}
    \caption{phrase-level skip-gram}
    \label{fig:1b}
\end{subfigure}
}
\caption{\small{Architecture of the skip-gram model.}}
\label{fig:skip-gram}
\vspace{-1em}
\end{figure}

Distributed word vector representations learned from large corpora of unlabeled data have been shown to be effective in
a variety of NLP tasks, such as POS tagging~\cite{collobert2011scratch}, parsing~\cite{chen2014fast,DurrettKlein2015}, and machine translation~\cite{devlin-EtAl:2014:P14-1,liu-EtAl:2014:P14-1,sutskever2014sequence,kalchbrenner2013recurrent}.
One of the most widely used approaches to learn these vector representations is the skip-gram model described in \namecite{Mikolov:2013} and \namecite{mikolov2013distributed}. 
The skip-gram model optimizes the probability of predicting words in the context given the current center word.  
Figure~\ref{fig:skip-gram} shows a standard skip-gram structure. 
\namecite{ling-EtAl:2015:NAACL-HLT} present a variation of skip-gram which captures the relative position of context words
by using a different weight matrix to connect the hidden layer to the context word at each relative position.

Word embeddings are usually hard to scale to larger units due to data sparsity. Recent work~\cite{Mitchell08vector-basedmodels,Baroni:2010,DBLP:journals/corr/abs-1003-4394,fyshe-EtAl:2015:NAACL-HLT} deals with this issue by constructing distributional representations for phrases from word embeddings.
\namecite{socher-EtAl:2013:ACL2013} use a recursive neural network
to learn weight matrices that capture compositionality.
However, the learning procedure is limited to the labeled Penn Treebank and the phrasal information
from the unlabeled data is not utilized.

\namecite{DBLP:journals/corr/LebretC15} 
propose a learning schedule which learns distributed vector representations for words and phrases jointly.
However, they haven't used the context information during the optimization procedure.
Recently \namecite{yu2015learning} propose a feature-rich compositional transformation (FCT) model
which learns weighted combination of word vectors to compose phrase vectors, where they mainly focus on bigram \textit{NP}s.

In this paper, we extend the skip-gram model to use the phrase structure in a large corpus to capture phrase compositionality and positional information
of both words and phrases. We jointly model words in the context of words and
phrases in the context of phrases. Additionally, we enforce a compositionality constraint 
on both the input and output phrase embedding spaces which indicates how
we build the distributed vector representations for phrases from their component word vectors.
Our results show that using phrase level context information provides gains in both word similarity and phrase similarity tasks.
Additionally, if we model phrase-level skip-gram over syntactic phrases, it would be helpful for syntactic tasks like syntactic analogy and 
dependency parsing.
\section{Skip-gram model}
\label{sec:skip}
The skip-gram model learns distributed vector representations for words by maximizing the probability of predicting
the context words given the current word. According to the \textit{word2vec} implementation of \namecite{mikolov2013distributed}, 
each input word $w$ is associated with a $d$-dimensional vector $v_w \in {\mathbb{R}}^{d}$ called the \textit{input embedding} and 
each context word $w_O$ is associated with a $d$-dimensional vector $v'_{w_O} \in {\mathbb{R}}^{d}$ called the \textit{output embedding}. $w, w_O$ are words from a vocabulary $V$
of size $W$. The probability of observing $w_O$ in the context of $w$ is modeled with a softmax function:
\begin{equation}
P(w_O | w) = \frac{\exp({v'_{w_O}}^T v_w)}{\sum_{i=1}^W \exp({v'_{w_i}}^T v_w)}
\end{equation}

The denominator of this function involves a summation over the whole vocabulary, which is impractical. One alternative 
to deal with the complexity issue is to sample several negative samples to avoid computing all the vocabulary. The objective
function after using negative sampling is:
\begin{equation}
\resizebox{.87\hsize}{!}{$\displaystyle
    E_w = \sum_{w \in s} (\log \sigma ({v'_{w_O}}^T v_w) + \sum_{i=1}^{K} \log \sigma (-{v'_{w_i}}^T v_w))
$}
\end{equation}
where $s$ is a chunked sentence. $w_i$, $i=1, 2, \ldots ,K$, are negative samples sampled from the following distribution:
\begin{equation}
P(w) = \frac{{\widetilde{P}(w)}^{\frac{3}{4}}}{Z}
\end{equation}
where $\widetilde{P}(w)$ is the unigram distribution of words and $Z$ is the normalization constant. The exponent $\frac{3}{4}$ is 
set empirically.

\section{Compositionality-aware skip-gram model}
\label{sec:compose}
To capture the way of composing phrase embeddings from distributed word vector representations, we extend the skip-gram model to include information from 
context of phrases and learn their compositionality from word vectors during the optimization procedure. Our phrase-level skip-gram structure is shown in
Figure~\ref{fig:1b}. 
\subsection{Phrase-level skip-gram model}
The word-level skip-gram model predicts the context words given the current word vector. Our approach further models the prediction of
context phrases given the vector representation of the current phrase vector (Figure~\ref{fig:1b}). Assume $v_p \in {\mathbb{R}}^d$ to be the $d$-dimensional 
input embedding for current phrase $p$ and $v'_{p_O}\in {\mathbb{R}}^d$ to be the output embedding for context phrase $p_O$. Using negative sampling, we model the phrase-level probability with:
\begin{equation}
\label{eq:phrase-skip}
\resizebox{.87\hsize}{!}{$\displaystyle
    E_p = \sum_{p\in s} (\log \sigma ({v'_{p_O}}^T v_p) + \sum_{i=1}^{N} \log \sigma (-{v'_{p_i}}^T v_p))
$}
\end{equation}
where $p_i$, $i= 1, 2, \ldots ,N$, are negative samples sampled according to the unigram probability of phrases raised to the same exponent $\frac{3}{4}$.

In this paper, we jointly model word-level skip-gram and phrase-level skip-gram for each sentence:
\begin{equation}
E = E_w + \beta E_p
\end{equation}
where $\beta > 0$ adjusts the relative importance of the word-level and the phrase-level skipgram.
\subsection{Compositional model}
Assume a phrase $p$ is composed of words $w_1, \ldots ,w_{n_p}$, where 
$n_p$ is the number of component words. The vector representation for $p$ is computed as:
\begin{equation}
    v_p = \Phi(\oplus( \sigma(v_{w_1}), \ldots ,\sigma(v_{w_i})))
\end{equation}
where $v_p$ is the vector representation for $p$. The function $\sigma$ is a component-wise manipulation over each dimension.
The symbol $\oplus$ is an operator over the component word vectors, which can be \textit{linear combination, summation, concatenation} etc. 
The mapping function $\Phi$ is a linear or non-linear manipulation over the resulting vector after the $\oplus$ operation. The same composition function is used to compute 
the output phrase embeddings $v'_{p_O}$ and $v'_{p_i}$, except that the component word vectors are $v'_{w_i}$ instead of $v_{w_i}$.

To show the effect of modeling phrase embeddings, we experiment with a composition function where $\oplus$ is \textit{linear combination} and $\Phi$ is passing the resulting matrix to the left of a weight vector $l^p = [l_1^p , l_2^p, \ldots, l_{n_p}^p]$
associated with each phrase $p$ showing how we combine the component word vectors. 
\begin{equation}
\label{eq:linear}
\resizebox{.55\hsize}{!}{$\displaystyle
v_p = [\sigma(v_{w_1}), \ldots, \sigma(v_{w_{n_p}})] \begin{bmatrix} l_1^p \\ \vdots\\ l_{n_p}^p \end{bmatrix}
$}
\end{equation}
where the function $\sigma$ is a component-wise power function over vector $v=[v_1, \ldots, v_n]$:
\begin{equation}
    \sigma(v) = [\phi(v_1), \ldots, \phi(v_n)]
\end{equation}
where $\phi(v_i)=\textit{sign}(v_i)|v_i|^{\alpha}$, $\alpha\geq 1$, is a power function over each dimension. This manipulation can be interpreted as adjusting dimensional values of word vectors to the phrase vector space.

Stochastic gradient ascent is used to update the word vectors. In equation~\ref{eq:phrase-skip}, for each word $w_j$ in $p'$, either context phrase $p_O$ or negative 
phrase sample $p_i$, the gradient is:
\begin{equation}
    \frac{\partial E_p}{\partial v'_{w_j}} = \triangledown \sigma(v'_{w_j}) (l_j^{p'} (y-\sigma({v'_{p'}}^T v_p)) v_p) 
\end{equation}
where $y=1$ for each word in $p_O$ and 0 for each word in $p_i$. $\triangledown \sigma(v'_{w_j})$ is a diagonal matrix where the $i$-th diagonal value is $\phi'({v'_{w_j}}_i)$.
For each word $w_j$ in the current phrase $p$, the gradient is:
\begin{equation}
\resizebox{.85\hsize}{!}{$\displaystyle
\frac{\partial E_p}{\partial v_{w_j}} = \triangledown \sigma(v_{w_j}) (l_j^{p} ((1-\sigma({v'_{p_O}}^T v_p)) v'_{p_O} + \sum_{i=1}^{N} (-\sigma({v'_{p_i}}^T v_p)) v'_{p_i})) 
$}
\end{equation}

\subsection{Output phrase embedding space}
Following \namecite{ling-EtAl:2015:NAACL-HLT} in 
using different output embeddings at each relative position to capture order information of context words (we call this \textit{positional} model), 
we use separate output embeddings to capture phrase-compositionality (we call this \textit{compositional} model). That is, we have a separate component 
word vector $v''$ to compose the phrase vectors in the context 
and the negative samples instead of using $v'$.
The intuition of this choice is that we don't want the compositionality information in the context or negative sample layer 
to be distorted by word-level updates.

We further extend the phrase-level skipgram to include the order information, which
uses different output word embeddings to compose phrases at each relative position. Phrases at the same relative position share the same
output embeddings (we call this model \textit{positional+compositional}).
Without loss of generality, we experiment with the composition
described in Equation~\ref{eq:linear}. The coefficients $l_i^p$s are set to be $\frac{1}{n_p}$.
\section{Experiments}
\label{sec:experiment}
We train the skip-gram model with negative sampling using \textit{word2vec} as our baseline. We used an April 2010 snapshot of the Wikipedia corpus~\cite{wiki2010}, which contains approximately 2 million articles and 990 million tokens. 
We remove all words that have a frequency less than 20 and use a context window size of 5 (5 words before and after the word occurrence). 
We set the number of negative samples to be 10 and the dimensionality of vectors to be 300. 
For phrase-level skip-gram, we also use a context window size of 5 (5 phrases before and after the phrase occurrence).

\subsection{Phrase compositionality}
\begin{table}
\centering
\begin{center}
\scalebox{0.8}{
\begin{tabular}{|l|l|} \hline
    & $\rho$ \\ \hline
    \namecite{Mitchell08vector-basedmodels} & 0.19 \\ \hline
   word2vec  & 0.23  \\ \hline
  compositional & 0.25 \\ \hline
 positional & 0.23 \\ \hline
    compositional+ positional & \textbf{0.26}\\ \hline
\end{tabular}}
\caption{\small{Spearman's correlation using different embeddings}}
\label{tab:phrase-compose}
\end{center}
\end{table}

We first evaluate the compositional model on the intransitive verb disambiguation dataset provided by~\namecite{Mitchell08vector-basedmodels}.
The dataset consists of pairs of subject and intransitive verb and a landmark intransitive verb is provided for each pair. For example,
this task requires one to identify when taking ``sale" as the subject, reference ``slump" and landmark ``decline" are close
to each other while ``slump" and landmark ``slouch" are not.
Each pair has multiple human ratings indicating how similar the pairs are.

We use the \textit{senna} toolkit
to extract the POS tag labels for each sentence and combine adjacent noun-verb pairs.
As the evaluation dataset is lemmatized, we also lemmatize the Wikipedia corpus to avoid sparsity. 
We evaluate the cosine similarity between the composed reference subject intransitive verb pair and its landmarks. Then we compute the Spearman's
correlation between the similarity scores and the human ratings. Table~\ref{tab:phrase-compose} shows the result. We can see that the compositions
of subject and intransitive verbs can be learned using the compositional model and the best performance is achieved using the joint model.
We use $\alpha=1$ and varying $\alpha$ does not change much of the performance.

\begin{table}
\centering
\begin{center}
\scalebox{0.70}{
\begin{tabular}{|l|l|l|l|l|} \hline
            & word353 & men & SYN & MIXED \\ \hline
 word2vec  & 0.722 & 0.758 & 69.9      & 77.8  \\ \hline
positional & 0.692 & 0.744 & 71.2 & 79.7 \\ \hline
compositional &  \textbf{0.735}  & \textbf{0.763}   & 70.6   & 79.2 \\ \hline
compositional+positional & 0.704 & 0.746 & \textbf{72.0} & \textbf{80.5} \\ \hline
\end{tabular}}
\caption{\small{Comparison of word-level tasks, including word similarity, analogy.}}
\label{tab:word-task}
\end{center}
\end{table}
\subsection{Word similarity and word analogy}
We also consider word similarity and analogy tasks for evaluating the quality of word embeddings.
Word similarity measures Spearman's correlation coefficient between the human scores
and the embeddings' cosine similarities for word pairs. Word analogy measures the accuracy on
syntactic and semantic analogy questions.

Here we extract phrases that contain syntax information, in the hope of learning additional syntactic 
information by modeling phrase-level skip-gram.
We use the \textit{senna} toolkit to identify the constituent chunks in each sentence. 
We run the chunker on 20 CPUs, and it takes less
than 2 hours to chunk the Wikipedia 2010 corpus we use.
The extracted phrases are labeled with \textit{NP, VP, PP}s etc. 

We evaluate similarity on two tasks, WordSim-353 and men, 
respectively containing 353 and 3000 word pairs. 
We use two word analogy datasets that we call SYN (8000 syntactic analogy questions) and MIXED (19544
syntactic and semantic analogy questions).

On the WordSim-353 task, \namecite{neelakantan-EtAl:2014:EMNLP2014} reported a Spearman's correlation of 0.709, while their word2vec baseline is 0.704.
From Table~\ref{tab:word-task}, we can see that positional information actually degrades the performance on the similarity task, while only adding compositional information
performs the best.
For syntactic and mixed analogy tasks which involves
syntax information, we can see that using phrase-level skip-gram and positional information both help and combining both gives the best performance.
 
\subsection{Dependency parsing}
\begin{table}
\centering
\begin{center}
\scalebox{0.7}{
\begin{tabular}{|l|l|l|l|l|} \hline
    & \multicolumn{2}{l|}{Dev} & \multicolumn{2}{l|}{Test} \\ \hline
   & UAS & LAS & UAS      & LAS  \\ \hline
   word2vec  & 92.21 & 90.83 &  91.91     & 90.54  \\ \hline
  compositional & 92.34  & 90.91  &  92.02     &  90.64 \\ \hline
 positional & 92.29 & 90.88 & 92.05 & 90.67  \\ \hline
    compositional+ positional & \textbf{92.39} & \textbf{90.91} & \textbf{92.19} & \textbf{90.82} \\ \hline
\end{tabular}}
\caption{\small{dependency parsing results on PTB using different embeddings}}
\label{tab:dependency-task}
\end{center}
\end{table}
We use the same preprocessing procedure as in the previous task. The evaluation on dependency parsing is performed on the English PTB, with the standard train, dev and
test splits with Stanford Dependencies. We use a neural network as described in \namecite{chen2014fast}. 
As we are using embeddings with a different dimensionality, we tune the hidden layer size and 
learning rate parameters for the neural network parser by grid search for each model and train for 15000 iterations. 
The other parameters are 
using the default settings. Evaluation is performed with the labeled (LAS) and unlabeled
(UAS) attachment scores. We run each parameter setting for 3 times and then average to prevent randomness.
We can see that by modeling phrase-level skip-gram over syntactic phrases, the performance on the dependency parsing task can be improved.
 
\section{Conclusion}
In this paper, we have presented a variation of skip-gram model which learns compositionality of phrase
embeddings. Our results show that modeling phrase-level co-occurrance 
and phrase compositionality helps improve word and phrase similarity tasks. 
If the phrases contain syntactic information, it would also help improve syntactic tasks.
As our compositionality function is very general, it would be interesting to see different variations and choices of composition function in different
applications.

%

\bibliography{bib/all}
\bibliographystyle{emnlp2016}

\end{document}